\documentclass[11pt,conference]{IEEEtran}

\usepackage[utf8]{inputenc}
\usepackage[T1]{fontenc}
\usepackage{amsmath}
\usepackage{amsfonts}
\usepackage{amssymb}
\usepackage{booktabs}
\usepackage{graphicx}
\usepackage{microtype}
\usepackage{nicefrac}
\usepackage{xcolor}
\usepackage{url}
\usepackage[numbers,sort&compress]{natbib}
\usepackage{tikz}
\usepackage{pgfplots}
\usepackage{hyperref}
\pgfplotsset{compat=1.18}

\graphicspath{{./}}
\hypersetup{
  colorlinks=true,
  linkcolor=black,
  citecolor=blue!55!black,
  urlcolor=blue!55!black,
  pdftitle={Beyond Target Scores: Measuring Off-Target Drift in Diffusion-Based Medical Image Editing},
  pdfauthor={Todd Y. Zhou},
  pdfkeywords={medical image editing, diffusion models, off-target drift, reward hacking, chest radiography}
}

\title{Beyond Target Scores: Measuring Off-Target Drift in Diffusion-Based Medical Image Editing}

\author{%
  \IEEEauthorblockN{Todd Y. Zhou}
  \IEEEauthorblockA{Harvard University}
}

\begin{document}

\maketitle
\pagestyle{plain}
\thispagestyle{plain}

\begin{abstract}
Diffusion models can now edit medical images in visually plausible ways, but the
standard evaluation question is too narrow: did the target score increase? In
clinical imaging, target findings are entangled with co-morbidities, acquisition
effects, and selection bias, so a model can appear successful by changing
correlated non-target findings rather than isolating the intended pathology. We
introduce \textbf{CIB-Med-1}, a trajectory-level benchmark for controlled
biomarker editing in chest radiography. CIB-Med-1 evaluates directional pleural
effusion editing through calibrated target progression, inversion rate, and
off-target semantic drift over 14 clinically motivated nuisance axes. The
benchmark exposes a reward-hacking failure mode in which diffusion editors
increase effusion scores while simultaneously altering parenchymal,
cardiomediastinal, pleural, chronic, or artifact-related findings. We further
present a constrained diffusion guidance baseline that optimizes target
progression subject to bounded off-target change. Across held-out radiographs,
the constrained editor preserves target progression ($\rho_{\mathrm{trend}}=0.88$
vs.\ $0.90$ for unconstrained guidance) while reducing median off-target drift
from $0.46$ to $0.20$ and 90th-percentile drift from $0.98$ to $0.33$. Drift
magnitude tracks empirical target--off-target association, supporting the view
that semantic instability is structured rather than incidental. A blinded human
validation probe with radiology trainees further shows stronger agreement with
intended progression orderings ($\tau=0.61$ vs.\ $0.29$ for Pix2Pix). These
results argue that medical image editing should be evaluated as trajectory-level
semantic control, not as endpoint score maximization.
\end{abstract}


\section{Introduction}

Diffusion-based generative models have enabled increasingly precise forms of
\emph{semantic image editing}, in which a model modifies a specified attribute of
an input image while attempting to preserve all other aspects of the scene
\citep{dhariwal2021diffusion, ho2020denoising, rombach2022high}. In medical
imaging, such capabilities are especially appealing. Clinicians and researchers
have long sought tools that can manipulate individual pathological factors (such
as disease severity, progression stage, or treatment response) while holding
other findings fixed, enabling counterfactual reasoning, data augmentation, and
controlled stress-testing of diagnostic systems
\citep{selvaraju2020gradcam, kermany2018identifying}.

Despite rapid empirical progress, \emph{directional medical image editing remains
poorly specified as a scientific problem}. Existing approaches are typically
evaluated using perceptual realism metrics or by measuring improvement in a
single target classifier score
\citep{gal2022image, kawar2023denoising}. These criteria are insufficient in
clinical settings, where pathological findings are statistically entangled
through comorbidity, shared imaging cues, and selection into observation
\citep{obermeyer2019dissecting, seyyed2020selection}. An editor that increases a
target disease score by inadvertently amplifying correlated, but clinically
distinct, findings may appear successful under standard metrics, while producing
images that violate the intended counterfactual interpretation.

This paper argues that the central challenge in medical image editing is
not realism, but identifiability. In observational clinical data, many findings
co-occur systematically, and learned representations often encode these
dependencies implicitly. As a result, optimizing for target improvement alone
permits a form of \emph{reward hacking}: models can exploit correlations in the
training distribution rather than manipulating the intended pathological factor
itself. Apparent gains in editing performance may therefore reflect better
shortcut exploitation, not improved semantic control
\citep{geirhos2020shortcut, amodei2016concrete}.

We propose that directional medical image editing should satisfy four core
criteria:
\begin{enumerate}
    \item \textbf{Monotone target progression:} edits should move consistently
    along a predefined clinical progression axis, such as increasing disease
    severity.
    \item \textbf{Off-target semantic stability:} non-target clinical findings
    should remain approximately invariant, up to a bounded tolerance.
    \item \textbf{Trajectory-level evaluation:} success should be assessed over
    the entire edit trajectory rather than at a single endpoint.
    \item \textbf{Reader alignment:} measured progression should correspond
    to clinically perceived severity under blinded human review.
\end{enumerate}
These requirements are not enforced by existing benchmarks or evaluation
protocols. In particular, evaluations that condition solely on target classifier
improvement conflate genuine progression with correlated semantic drift, making
it impossible to distinguish independent manipulation from distribution-following
behavior.

\paragraph{Contributions.}
This paper makes four contributions. First, we introduce
\textbf{CIB-Med-1}, a benchmark for \emph{Controlled Incremental Biomarker}
editing in chest radiography that makes failures of semantic identifiability
explicit and measurable. The benchmark anchors edits at real chest radiographs
and evaluates generated trajectories using clinically grounded semantic
coordinates derived from a frozen radiology classifier. Second, CIB-Med-1
measures not only target progression but also \emph{off-target drift} across a
curated set of non-target findings, exposing reward-hacking behaviors that remain
invisible under standard metrics. Third, we show that diffusion-based editors can
achieve apparent target improvement by inducing correlated changes in
parenchymal, cardiomediastinal, pleural, chronic, and artifact-related findings
rather than by isolating the intended pathology. Fourth, we present a constrained
diffusion guidance baseline that approximately optimizes target progression
subject to bounded off-target semantic change. Through ablation studies and
semi-synthetic stress tests, we identify which semantic constraints are most
critical for preventing drift and characterize the limits of independent
manipulation under correlated data.

\paragraph{Scope.}
This work does not claim causal disentanglement from observational data, nor does
it aim to simulate clinically faithful disease progression. Rather, our goal is
to make explicit what directional editing methods are optimizing and to provide
tools for evaluating whether progress along a desired axis is achieved
independently or only through correlated shortcuts. By reframing medical image
editing as a problem of semantic identifiability and trajectory-level control, we
aim to enable more principled development and evaluation of generative models in
high-stakes clinical domains. As an initial human validation, a blinded study
with two radiology trainees found that our constrained editor achieved a Kendall
rank correlation of $\tau=0.61$ with perceived progression order, compared with
$\tau=0.47$ for unconstrained diffusion and $\tau=0.29$ for Pix2Pix.

\begin{figure}[t]
    \centering
    \includegraphics[width=\columnwidth]{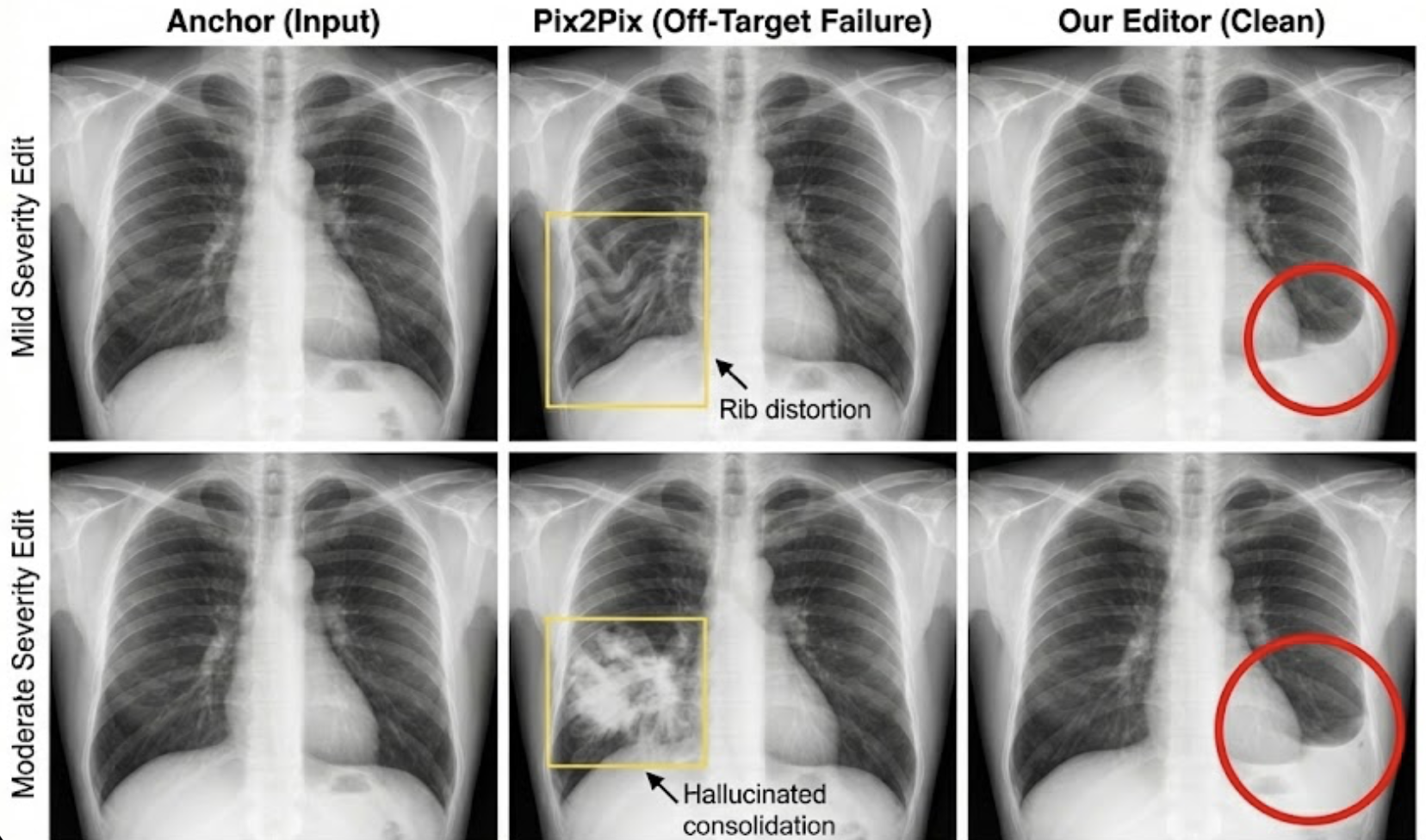}
    \caption{\textbf{Target-only evaluation can reward the wrong behavior.}
    A medical image editor may increase the intended effusion score while also
    changing correlated findings. CIB-Med-1 evaluates the full trajectory and
    explicitly measures these off-target changes.}
    \label{fig:motivation}
\end{figure}

\section{Problem Setup and Semantic Coordinates}
\label{sec:setup}

We study \emph{directional medical image editing}: given an input image, an editor
should move the image along a specified clinical axis (the \emph{target}) while
preserving other clinically relevant attributes (the \emph{off-target} axes).
Our goal is to make this objective explicit and measurable at the level of
\emph{edit trajectories}, thereby distinguishing genuine semantic control from
distribution-following behavior that exploits dataset correlations.

\subsection{Anchors, Trajectories, and Editors}
\label{sec:trajectories}

Let $x^0 \in \mathbb{R}^{H \times W}$ denote an \emph{anchor} chest radiograph
(CXR). An editor induces a trajectory $\{x^t\}_{t=0}^T$ where each
$x^{t+1}$ is produced by an editing operator $\mathcal{E}$ (e.g., a guided
diffusion sampler) applied to the current state:
\begin{equation}
x^{t+1} = \mathcal{E}(x^t;\, \eta_t), \qquad t=0,\dots,T-1
\end{equation}
where $\eta_t$ collects step-specific randomness and/or guidance parameters.
A successful directional editor should yield trajectories that exhibit
(i) monotone progression in the target coordinate and (ii) stability in
off-target coordinates, as defined below.

\subsection{Evaluator and Clinical Semantic Coordinates}
\label{sec:coordinates}

A central challenge in medical editing is that the desired semantics (disease
severity, co-occurring findings) are not directly observable as ground-truth
continuous factors. We therefore evaluate trajectories using a fixed,
out-of-training \emph{radiology evaluator} that maps images to clinically
interpretable multi-label scores. While $F$ is a proxy, our human validation
(Section~\ref{sec:results_validation7}) suggests that optimizing for $F$ under
semantic constraints yields trajectories that radiology readers perceive as more
clinically ordered than unconstrained optimization. Let
$F:\mathbb{R}^{H\times W}\rightarrow \mathbb{R}^{|\mathcal{L}|}$ be a frozen
multi-label CXR classifier producing logits
$\ell(x) = (\ell_k(x))_{k\in\mathcal{L}}$ over a label set $\mathcal{L}$
(e.g., standard CXR findings). We define score coordinates using logistic
probabilities
\begin{equation}
v_k(x) := \sigma(\ell_k(x)) \in (0,1), \qquad k\in\mathcal{L},
\label{eq:vk_def}
\end{equation}
where $\sigma$ is the sigmoid. We emphasize that $F$ is used \emph{only} for
evaluation and guidance; it is not trained on generated samples. This evaluator
choice makes our criteria falsifiable and yields a transparent decomposition
into target vs.\ off-target axes.

\paragraph{Target coordinate $p_{\mathrm{eff}}(x)$.}
We focus on pleural effusion as a canonical progression axis in CXR editing.
Let $\ell_{\mathrm{eff}}(x)$ denote the evaluator logit for pleural effusion.
Because diffusion guidance is sensitive to score scaling and because raw
classifier probabilities are often miscalibrated in clinical data, we define a
monotone, calibrated target coordinate
\begin{equation}
p_{\mathrm{eff}}(x) := g(\ell_{\mathrm{eff}}(x)) \in [0,1],
\label{eq:peff_def}
\end{equation}
where $g$ is an isotonic regression map fit on a held-out labeled set. The map
$g$ is order-preserving by construction, so $p_{\mathrm{eff}}$ provides an
interpretable one-dimensional progression coordinate while remaining compatible
with gradient-based guidance via $\ell_{\mathrm{eff}}$. In short:
$\ell_{\mathrm{eff}}$ provides a differentiable guidance signal, while
$p_{\mathrm{eff}}$ provides a calibrated reporting coordinate.

\paragraph{Off-target coordinate set $\mathcal{K}$ and $v_k$.}
We define a curated set of \emph{off-target findings} $\mathcal{K}\subset
\mathcal{L}\setminus\{\mathrm{eff}\}$ whose scores should remain stable as
effusion changes. Concretely, we choose a compact set covering (i) common
parenchymal findings, (ii) cardiomediastinal findings, (iii) pleural findings
distinct from effusion, and (iv) artifact/intervention ``canaries'':
\begin{equation}
\begin{aligned}
\mathcal{K}=\{&\texttt{Atelectasis},\ \texttt{Consolidation},\\
&\texttt{Pneumonia},\ \texttt{Edema},\ \texttt{Lung~Opacity},\\
&\texttt{Cardiomegaly},\\
&\texttt{Enlarged~Cardiomediastinum},\\
&\texttt{Pneumothorax},\ \texttt{Pleural~Thickening},\\
&\texttt{Pleural~Other},\ \texttt{Fibrosis},\ \texttt{Emphysema},\\
&\texttt{Support~Devices},\ \texttt{Fracture}\}.
\end{aligned}
\label{eq:K_def}
\end{equation}
For each $k\in\mathcal{K}$, we define the off-target semantic coordinate
$v_k(x)$ as in Eq.~\eqref{eq:vk_def}. Thus, off-target findings are fixed before
evaluation as a clinically motivated subset of the evaluator label space, and
each $v_k$ is the evaluator score for label $k$.

\subsection{Monotone Progression and Off-Target Stability}
\label{sec:desiderata}

Given a trajectory $\{x^t\}_{t=0}^T$ from anchor $x^0$, we require:

\paragraph{(D1) Monotone target progression.}
Edits should move consistently in the target direction. We operationalize this
using trajectory-level monotonicity (and later report inversion-based metrics):
\begin{equation}
p_{\mathrm{eff}}(x^{t+1}) \ge p_{\mathrm{eff}}(x^{t}) \qquad \forall t.
\label{eq:monotone}
\end{equation}

\paragraph{(D2) Off-target semantic stability.}
Edits should not create or erase non-target findings. We operationalize stability
relative to the anchor:
\begin{equation}
v_k(x^{t}) \approx v_k(x^0) \quad \forall k\in\mathcal{K},\ \forall t.
\label{eq:stability}
\end{equation}
In practice we quantify stability through an off-target drift functional defined
in Section~\ref{sec:metrics}. Crucially, the stability requirement is stated in
the evaluator's semantic coordinate system, making it possible to detect when an
editor improves the target score by changing correlated findings.

\paragraph{(D3) Trajectory-level evaluation.}
Because reward hacking can occur gradually (e.g., by slowly increasing global
opacity or altering texture statistics), we evaluate entire trajectories rather
than a single endpoint. This is essential for distinguishing independent control
of $p_{\mathrm{eff}}$ from distribution-following drift in $v(x)$.

\subsection{Why Off-Target Drift Arises: Correlation, Entanglement, and MNAR}
\label{sec:drift_origin}

A central observation motivating this work is that off-target drift is not a
rare failure mode. It is a predictable consequence of how clinical data are
generated and how models learn from it.

\paragraph{Correlation and representation entanglement.}
In real-world hospital data, pleural effusion frequently co-occurs with other
findings (e.g., edema, cardiomegaly, atelectasis, parenchymal opacity). Moreover,
the visual cues that support one diagnosis may overlap with cues for others (e.g.,
global density shifts, basilar opacities). Therefore, a guidance signal that
increases pleural effusion can also increase correlated $v_k$ unless the editor
is explicitly constrained. In Section~\ref{sec:analysis}, we quantify this by
measuring empirical associations between $p_{\mathrm{eff}}$ and each $v_k$ on the
anchor distribution and show that the largest drift dimensions often coincide
with the most strongly associated labels. This alignment suggests that semantic
instability is a manifestation of learned representational entanglement: the
guidance field inherits coupled structure from the training distribution and may
not separate genuine physiological co-morbidity from spurious, selection-biased
shortcuts.

\paragraph{Spurious correlations and clinically structured confounding.}
Some off-target changes are clinically plausible co-morbidity, while others may
reflect shortcut learning or dataset-induced confounding. For example, pleural
effusion can be associated with infectious or chronic processes in particular
subpopulations, and radiographic patterns linked to conditions such as
tuberculosis may co-vary with effusion prevalence in specific cohorts. If an
editor increases effusion by also amplifying such correlated patterns, the edit
trajectory may look ``successful'' under target-only metrics while violating the
intended counterfactual semantics. This motivates our emphasis on off-target
drift as both an evaluation axis and a diagnostic signal about entanglement.

\paragraph{Selection and non-missing-at-random (MNAR) mechanisms.}
Clinical imaging datasets are observational: the probability that an image is
acquired and labeled depends on unobserved severity, comorbidities, care
pathways, and institutional practices. As a result, observed associations between
findings can be induced or amplified by \emph{selection into observation}. This
violates missing-at-random assumptions and implies that empirical correlations
among labels may reflect both physiology and selection effects. Consequently, an
editor optimized against observational signals can inherit dataset-induced
dependencies, producing images that are statistically consistent with the
training distribution yet unsuitable as counterfactual interventions. We address
this by (i) explicitly measuring drift across multiple nuisance axes,
(ii) performing semi-synthetic stress tests with known ground-truth perturbations
(Section~\ref{sec:synth}), and (iii) reporting stratified analyses where possible.

\subsection{Distinguishing between Causal Simulation and Independent Control}
A common critique of penalizing off-target drift is that doing so may discourage
physiological realism, since findings such as effusion and atelectasis can be
linked in vivo. We therefore distinguish two goals: \emph{simulation}, which
models the joint observational distribution $P(y\mid x)$, and \emph{control},
which asks whether a model can move along one specified semantic axis while
holding measured nuisance axes approximately fixed. CIB-Med-1 evaluates the
latter. For model debugging, robustness testing, and counterfactual analysis, the
relevant question is: what changes when the intended effusion coordinate changes,
and what else changes with it? A model that can increase the target score only by
hallucinating downstream or correlated findings, even biologically plausible
ones, fails the test of independent controllability. CIB-Med-1 is designed to
detect this failure, making it a stress test for representation learning rather
than a claim of biological simulation.

\subsection{Implication: Why ``Better Models'' Are Not Sufficient}
\label{sec:why_not_better_models}

The above considerations imply that optimizing the wrong objective can produce
apparent progress even as semantic control deteriorates. If evaluation rewards
target-score improvement without penalizing correlated drift, then editors can
``win'' by exploiting representational or dataset correlations rather than by
independently manipulating the intended factor. This is precisely the failure
mode we refer to as reward hacking in directional medical editing. Therefore, the
appropriate response is not merely to build a larger or higher-fidelity
generator, but to (i) define a use-case-aligned objective that makes drift
measurable and (ii) develop methods that optimize this objective without relying
on shortcuts. CIB-Med-1 provides the benchmark and coordinate system required for
this shift; our constrained guidance baseline illustrates one concrete approach.

\section{Trajectory-Level Metrics for Directional Editing}
\label{sec:metrics}

This section formalizes the evaluation criteria underlying CIB-Med-1. Our goal is
to distinguish independent semantic progression along a target clinical
axis from correlated drift along non-target axes. We therefore define
metrics that operate on entire edit trajectories and are expressed in the
clinical semantic coordinates introduced in Section~\ref{sec:setup}.

\subsection{Why Existing Metrics Are Insufficient}
\label{sec:why_metrics_fail}

Most prior work on image editing evaluates success using one of two criteria:
(i) perceptual realism, often through FID or CLIP-based similarity, or
(ii) improvement in a single target classifier score
\citep{heusel2017gans,radford2021learning}. Neither criterion is adequate for
directional medical editing.

Perceptual metrics are agnostic to clinical semantics and cannot distinguish
clinically meaningful changes from imperceptible or spurious texture shifts.
Target-only metrics, meanwhile, implicitly assume that the target attribute can
be manipulated independently. In observational clinical data, this assumption
fails: target findings are often statistically entangled with other conditions
through co-morbidity, shared visual cues, and selection into observation. As a
result, editors can increase a target score by moving along correlated nuisance
axes (a form of reward hacking that remains invisible under target-only
evaluation).

This motivates trajectory-level metrics that jointly assess (i) monotone target
progression and (ii) off-target semantic stability.

\subsection{Monotone Target Progression}
\label{sec:monotone_metrics}

Let $\{x^t\}_{t=0}^T$ be an edit trajectory from anchor $x^0$, and let
$p_{\mathrm{eff}}(x)$ be the calibrated effusion coordinate defined in
Eq.~\eqref{eq:peff_def}. We evaluate monotonicity using two complementary metrics.

\paragraph{Trend correlation.}
We compute the Spearman rank correlation between the step index $t$ and
$p_{\mathrm{eff}}(x^t)$:
\begin{equation}
\rho_{\mathrm{trend}} :=
\mathrm{corr}_{\mathrm{Spearman}}\bigl(t,\ p_{\mathrm{eff}}(x^t)\bigr).
\end{equation}
A value of $\rho_{\mathrm{trend}} \approx 1$ indicates consistent progression,
while lower values indicate stagnation or reversals.

\paragraph{Inversion rate.}
We define the inversion rate as the fraction of steps that violate monotonicity:
\begin{equation}
\text{InvRate} :=
\frac{1}{T}\sum_{t=0}^{T-1}
\mathbb{I}\!\left[p_{\mathrm{eff}}(x^{t+1}) < p_{\mathrm{eff}}(x^{t})\right].
\end{equation}
This metric penalizes even brief regressions and is sensitive to local failures
that may be obscured by global trend measures. Together, $\rho_{\mathrm{trend}}$ and $\mathrm{InvRate}$ capture both global and
local adherence to the desired progression direction.

\subsection{Off-Target Semantic Drift}
\label{sec:drift_metrics}

To quantify deviations along non-target axes, we measure drift relative to the
anchor $x^0$ in the off-target coordinates $v_k(x)$ defined in
Eq.~\eqref{eq:vk_def}.

\paragraph{Per-label drift.}
For each off-target finding $k\in\mathcal{K}$, we define the trajectory-averaged
absolute drift:
\begin{equation}
D_k :=
\frac{1}{T}\sum_{t=1}^{T} \left| v_k(x^t) - v_k(x^0) \right|.
\label{eq:Dk}
\end{equation}
This quantity measures how much the edit trajectory deviates from the anchor
along a specific semantic axis.

\paragraph{Aggregate off-target drift.}
Because different findings may exhibit different noise levels and baseline
prevalence, we aggregate per-label drift using robust statistics:
\begin{align}
D_{\mathrm{off}} &:= \mathrm{median}_{k\in\mathcal{K}}\, D_k, \\
D_{\mathrm{off}}^{(90)} &:= \mathrm{quantile}_{0.9}\bigl(\{D_k\}_{k\in\mathcal{K}}\bigr).
\end{align}
The median captures typical semantic deviation, while the upper quantile reflects
upper-tail collateral change, which is often most relevant in clinical settings.

\subsection{Correlation-Aware Interpretation of Drift}
\label{sec:drift_interpretation}

Off-target drift is not necessarily arbitrary noise. In clinical data, some
off-target findings are empirically correlated with the target condition.
Therefore, we explicitly analyze the relationship between drift magnitude and
label association. Let $\mathrm{Assoc}(k)$ denote an empirical association measure between
$p_{\mathrm{eff}}$ and $v_k$ computed on the anchor distribution (e.g., Spearman
correlation or conditional mutual information). In Section~\ref{sec:analysis}, we
show that findings with larger $\mathrm{Assoc}(k)$ tend to exhibit larger $D_k$
under unconstrained editing. This pattern suggests that drift often reflects
learned coupling structure (arising from a mixture of physiological co-morbidity,
spurious correlations, and selection effects) rather than random artifacts. This
observation allows off-target drift to be interpreted not only
as a failure mode, but also as a diagnostic signal revealing which semantic axes
are entangled with the target in the learned representation.

\subsection{Ablation-Based Importance of Off-Target Constraints}
\label{sec:ablation_metrics}

To assess which off-target findings are most critical for preventing drift, we
define an ablation-based importance analysis that operates directly on the drift
metrics above.

\paragraph{Leave-one-out ablation.}
For each $k\in\mathcal{K}$, we recompute drift metrics after removing $k$ from the
off-target set used by the editor or constraint mechanism. Let
$D_{\mathrm{off}}^{(-k)}$ denote the resulting aggregate drift. We define the
importance of $k$ as
\begin{equation}
\Delta_k := D_{\mathrm{off}}^{(-k)} - D_{\mathrm{off}}.
\end{equation}
A larger $\Delta_k$ indicates that constraining $v_k$ plays a substantial role in
preventing collateral semantic change.

\paragraph{Grouped ablations.}
We further group off-target findings into clinically meaningful categories
(parenchymal, pleural-non-effusion, cardiomediastinal, chronic, artifact) and
repeat the ablation analysis at the group level. This yields interpretable
insights into which classes of findings dominate drift control.

\subsection{Relation to MNAR and Selection Effects}
\label{sec:mnar_metrics}

Because clinical imaging data are not missing at random, observed correlations
between target and off-target findings may be induced or amplified by selection
into imaging and labeling. Our metrics do not assume that $p_{\mathrm{eff}}$ and
$v_k$ correspond to causally independent factors. Instead, they are explicitly
designed to surface violations of independent manipulability. By reporting drift
across multiple nuisance axes and relating it to empirical
associations, we make transparent when an editor's apparent success relies on
distribution-following behavior rather than independent semantic control. This
distinction is essential for evaluating editing methods intended for
counterfactual analysis, robustness testing, or any further clinical simulation.

CIB-Med-1 therefore evaluates directional medical image editing as a
trajectory-level control problem: successful editors must make monotone progress
along the target coordinate while limiting collateral movement along measured
off-target axes. In the following sections, we use these metrics to compare
editing methods, analyze sources of drift, and evaluate constrained guidance
strategies that mitigate collateral semantic change.

\section{Editing Methods: Constrained Diffusion Guidance}
\label{sec:method}

The metrics defined in Section~\ref{sec:metrics} expose a fundamental limitation
of existing image editing methods: optimizing for target improvement alone does
not enforce independent semantic control. In this section, we present a
constrained diffusion guidance method designed explicitly to satisfy the
trajectory-level requirement of CIB-Med-1. While our primary contribution is
evaluative, this method serves as a principled baseline and illustrates how
reward hacking can be mitigated when the objective is aligned with the use case.

\subsection{Classifier-Guided Diffusion Editing}
\label{sec:classifier_guidance}

We consider diffusion-based editors that generate samples by iteratively denoising
a latent variable $x_t$ toward an image $x_0$, guided by gradients of a semantic
objective. In standard classifier-guided diffusion, guidance toward a target
attribute is implemented by modifying the reverse diffusion update using the
gradient of a classifier log-probability \citep{dhariwal2021diffusion}.
Let $\ell_{\mathrm{eff}}(x)$ denote the evaluator logit for pleural effusion.
Unconstrained target guidance takes the form
\begin{equation}
\nabla_x \mathcal{G}_{\mathrm{target}}(x)
=
\nabla_x \ell_{\mathrm{eff}}(x),
\label{eq:target_guidance}
\end{equation}
which biases the sampling process toward images with higher effusion scores.
While effective at increasing $p_{\mathrm{eff}}$, this update implicitly assumes
that $\ell_{\mathrm{eff}}$ can be increased without altering other clinically
relevant attributes (an assumption violated in correlated clinical data).

\subsection{Constrained Objective: Progression with Bounded Drift}
\label{sec:constrained_objective}

To formalize the criterion of Section~\ref{sec:setup}, we conceptualize each edit
step as approximately solving a constrained optimization problem:
\begin{equation}
\max_{\Delta}\quad p_{\mathrm{eff}}(x + \Delta)
\quad \text{s.t.} \quad
\|v(x + \Delta) - v(x^0)\|_1 \le \epsilon,
\label{eq:constrained_problem}
\end{equation}
where $v(x) = (v_k(x))_{k\in\mathcal{K}}$ denotes the vector of off-target semantic
coordinates and $\epsilon$ is a tolerance parameter. This formulation makes
explicit that the goal is not merely to increase the target score, but to do so
without inducing unbounded collateral semantic change. Solving
Eq.~\eqref{eq:constrained_problem} exactly is intractable within a diffusion
sampler. We therefore use a first-order approximation that can be
implemented via gradient guidance.

\subsection{Signed Off-Target Barrier Guidance}
\label{sec:barrier_guidance}

We implement the constraint in Eq.~\eqref{eq:constrained_problem} using a
directional barrier that penalizes movement away from the anchor $x^0$ along each
off-target axis. The resulting guidance field is
\begin{equation}
\begin{aligned}
\nabla_x \mathcal{G}(x)
={}&\nabla_x \ell_{\mathrm{eff}}(x)\\
&-\lambda\sum_{k\in\mathcal{K}}w_k(x^0)\\[-2pt]
&\quad\times\operatorname{sgn}\!\left(v_k(x)-v_k(x^0)\right)
\nabla_x \ell_k(x),
\end{aligned}
\label{eq:full_guidance}
\end{equation}
where $\lambda > 0$ controls the strength of the constraint and
$w_k(x^0)$ are anchor-dependent weights defined below.
The \textbf{signed term} ensures that once $v_k(x)$ deviates from its anchor value, the
guidance applies force in the direction that reduces further deviation. This
construction differs from symmetric penalties (e.g., squared deviations) in that
it preserves small stochastic fluctuations while preventing systematic drift
over the trajectory.

\subsection{Anchor-Adaptive Off-Target Weights}
\label{sec:weights}

Uniform constraints across all off-target findings are suboptimal: some findings
are already present in the anchor image, while others are absent. Some axes are
empirically unstable or highly correlated with the target, while others are not.
We therefore define anchor-adaptive weights
\begin{equation}
w_k(x^0)
=
1
+
\alpha \, \mathrm{Var}\!\left(v_k \mid p_{\mathrm{eff}}\in\mathcal{N}(p_{\mathrm{eff}}(x^0))\right)
+
\beta \, v_k(x^0),
\label{eq:weights_def}
\end{equation}
where $\mathcal{N}(\cdot)$ denotes a local neighborhood in target space. The
variance term upweights off-target axes that exhibit high variability among
clinically similar anchors, which may indicate spurious correlations or
instability. The anchor-value term upweights findings already present in the
image, discouraging their inadvertent amplification or erasure. Together, these
terms bias the editor toward preserving the anchor's semantic profile while still
allowing target progression.

\subsection{Relation to Reward Hacking}
\label{sec:reward_hacking}

Unconstrained guidance in Eq.~\eqref{eq:target_guidance} permits reward hacking:
the editor can increase $p_{\mathrm{eff}}$ by moving along correlated off-target
axes that also raise the effusion score. In contrast, the constrained guidance in
Eq.~\eqref{eq:full_guidance} explicitly penalizes such movement, forcing the
editor to seek directions in image space that preferentially affect the target
coordinate. When such directions do not exist, or are weak due to representational
entanglement, the editor exhibits slower progression or saturates, revealing a
limit of independent manipulability rather than exploiting shortcuts.

\subsection{Ablations and Diagnostic Use}
\label{sec:method_ablation}

The structure of Eq.~\eqref{eq:full_guidance} enables systematic ablations that
directly connect method behavior to the metrics of Section~\ref{sec:metrics}. In
particular, removing a single off-target term $k$ from the sum corresponds to the
leave-one-out ablation used to compute $\Delta_k$ in
Section~\ref{sec:ablation_metrics}. Grouped ablations correspond to removing
entire subsets of terms. These experiments allow us to identify which semantic
axes are most responsible for preventing drift and to relate their importance to
empirical target-off-target correlations.

\subsection{Scope and Limitations}
\label{sec:method_scope}

We emphasize that constrained guidance does not guarantee causal disentanglement
of clinical factors. Because the evaluator is trained on observational data with
non-missing-at-random structure, gradients of $\ell_k$ may reflect both
physiological signals and dataset-induced shortcuts. The role of the constrained
editor is therefore not to ``correct'' these biases, but to make their effects
visible under trajectory-level evaluation. In this sense, the method functions as
a diagnostic tool aligned with the benchmark, rather than as a definitive
solution to disentanglement in clinical imaging.

\section{Experimental Setup and Baselines}
\label{sec:experiments}

This section describes the experimental protocol used to evaluate directional
editing methods under the CIB-Med-1 benchmark. Our goal is to ensure that all
comparisons are reproducible, evaluator-consistent, and aligned with the
trajectory-level metrics defined in Section~\ref{sec:metrics}.

\subsection{Datasets and Anchor Selection}
\label{sec:anchors}

We construct edit trajectories starting from real chest radiographs drawn from a
held-out clinical dataset of 5,000 images that is disjoint from any data used to
train the generative models. From this pool, we select a set of \emph{anchors}
$\{x^0_i\}_{i=1}^N$ satisfying three criteria. First, each anchor exhibits a
non-saturated target score $p_{\mathrm{eff}}(x^0_i)\in(0.1,0.9)$, ensuring that
both upward progression and stability are meaningfully testable. Second, each
anchor has well-defined evaluator outputs across the off-target set
$\mathcal{K}$, with no missing or degenerate scores. Third, anchors are sampled
across a broad range of baseline clinical profiles (e.g., low vs.\ high
cardiomegaly, presence vs.\ absence of parenchymal findings) to probe
heterogeneity in editing behavior.

This selection procedure avoids trivial cases where the target coordinate is
already maximized and mitigates selection bias toward visually simple anchors.
Importantly, anchor selection is performed \emph{without reference to the editing
methods}, ensuring that no method is advantaged by anchor choice.

\subsection{Trajectory Generation Protocol}
\label{sec:trajectory_protocol}

For each anchor $x^0$, we generate a trajectory
$\{x^t\}_{t=0}^{T}$ using each editing method under evaluation. All editing
methods utilize a Stable Diffusion v2.1 backbone fine-tuned on the CheXpert
training set to ensure medical domain competence
\citep{rombach2022high,irvin2019chexpert}. By using the same underlying weights
for all methods, we isolate the effect of the guidance strategy from the
generative capacity of the model. Unless otherwise stated, we set
$T=20$,\footnote{Empirically $T=20$ with our scheduler provided enough
trajectory resolution for semantic editing without obvious visual degradation in
pilot sweeps.} which balances trajectory resolution with computational cost. At
each step, guidance parameters are held fixed across anchors for a given
method. Hyperparameters are selected once on a disjoint calibration split, then
frozen; no method is tuned per-anchor or against the held-out evaluation set.
Stochasticity in the diffusion process is preserved and all reported metrics are
averaged over multiple random seeds.

\subsection{Evaluator and Calibration Details}
\label{sec:evaluator_setup}

All metrics are computed using the frozen radiology evaluator $F$ introduced in
Section~\ref{sec:setup}. The evaluator is not trained or fine-tuned on generated
images and is used consistently across all methods. The isotonic calibration map
$g$ defining $p_{\mathrm{eff}}$ is fit on a labeled
validation split disjoint from both the anchor set and the generative model
training data. Calibration is performed once and reused across all experiments.
This separation prevents leakage between evaluation and editing and ensures that
$p_{\mathrm{eff}}$ serves as a stable semantic coordinate.

\begin{table}[t]
\centering
\caption{CIB-Med-1 evaluation protocol. All settings are fixed across methods.}
\label{tab:protocol_summary}
\footnotesize
\setlength{\tabcolsep}{3.5pt}
\resizebox{\columnwidth}{!}{%
\begin{tabular}{ll}
\toprule
\textbf{Component} & \textbf{Specification} \\
\midrule
Held-out image pool & 5{,}000 real chest radiographs \\
Anchor selection & $p_{\mathrm{eff}}(x^0) \in (0.1, 0.9)$, valid scores for all $k \in \mathcal{K}$ \\
Number of anchors ($N$) & 240 \\
Diffusion backbone & Stable Diffusion v2.1 (CheXpert fine-tuned) \\
Trajectory length ($T$) & 20 steps \\
Random seeds per anchor & 5 \\
Target evaluator & Frozen radiology classifier $F$ \\
Calibration method & Isotonic regression on disjoint validation split \\
Off-target set size ($|\mathcal{K}|$) & 14 \\
Hyperparameter selection & Disjoint calibration split; frozen for evaluation \\
\bottomrule
\end{tabular}
}
\end{table}

Every numerical result is computed directly from completed per-anchor,
per-seed evaluations under this fixed protocol.

\subsection{Baselines}
\label{sec:baselines}

We compare the constrained diffusion guidance method of
Section~\ref{sec:method} against a set of representative baselines that reflect
common practices in image editing. \textbf{Unconstrained guidance} is the
clinical baseline: standard classifier-guided diffusion using the effusion logit
$\ell_{\mathrm{eff}}$ as the sole guidance signal
(Eq.~\eqref{eq:target_guidance}). This baseline isolates the effect of optimizing
the target coordinate without off-target constraints.
\textbf{Prompt-based diffusion editing} conditions generation on a textual
description corresponding to
increased effusion severity. While widely used in natural-image editing, such
methods do not explicitly encode clinical off-target stability and therefore
serve as a useful contrast.
\textbf{Instruction-guided image-to-image editing} uses
InstructPix2Pix~\citep{brooks2023instructpix2pix}, conditioned on the source
radiograph and a higher-effusion instruction. We abbreviate this baseline as
Pix2Pix in compact figures and tables. The learned mapping can absorb dataset
correlations and therefore provides a strong test of off-target preservation.
All baselines operate on the same anchors and are evaluated using identical
metrics. No baseline is tuned using CIB-Med-1 metrics, reflecting realistic
usage scenarios and preventing metric overfitting.

\subsection{Ablation Protocol}
\label{sec:ablation_protocol}

To identify which off-target constraints are most critical for preventing drift,
we perform systematic ablations aligned with the definitions in
Section~\ref{sec:ablation_metrics}. In the \emph{leave-one-out} setting, each
$k\in\mathcal{K}$ is removed in turn from the constrained guidance term. In the
\emph{grouped} setting, clinically coherent groups such as parenchymal or
cardiomediastinal findings are removed together. In the \emph{weight} setting,
anchor-adaptive weights $w_k(x^0)$ are replaced with uniform weights to assess
the contribution of adaptivity. For each ablation, we recompute all
trajectory-level metrics, ensuring that conclusions about constraint importance
are driven by metric-defined effects rather than qualitative inspection.

\section{Results}
\label{sec:results}

We now evaluate directional editing methods under the CIB-Med-1 benchmark using
the trajectory-level metrics defined in Section~\ref{sec:metrics}. Four findings
stand out. First, target-only methods achieve strong apparent progression, but
they do so with substantial off-target drift. Second, our constrained editor
retains comparable target progression ($\rho_{\mathrm{trend}}=0.88$ vs.\ $0.90$
for unconstrained guidance) while reducing median drift by 57\% and
90th-percentile drift by 66\%. Third, drift is structured rather than random:
the largest drift dimensions align with empirical target--off-target
associations in the anchor distribution. Fourth, lower drift corresponds to
better human-perceived progression ordering in a blinded radiology-trainee
validation probe.

\subsection{Target Progression Under Unconstrained Editing}
\label{sec:results_target}

We begin by evaluating monotone target progression. Across all anchors,
unconstrained classifier-guided diffusion reliably increases the effusion
coordinate $p_{\mathrm{eff}}$ and achieves trend correlation
$\rho_{\mathrm{trend}}=0.90$ with an inversion rate of $0.05$. Prompt-based and
image-to-image baselines also show strong average progression, though with
greater variability across anchors.
Under target-only metrics, these results would suggest that all methods are
effective at directional editing. However, as we show below, this apparent
success masks substantial collateral semantic change.

\subsection{Off-Target Drift Reveals Reward Hacking}
\label{sec:results_drift}

Despite strong target progression, unconstrained classifier guidance exhibits large
aggregate drift $D_{\mathrm{off}}$ and upper-tail drift
$D_{\mathrm{off}}^{(90)}$. In qualitative inspections, this drift corresponds to
systematic changes in parenchymal opacity, cardiomediastinal silhouette, and
pleural texture, features that are known to correlate with pleural effusion in
observational datasets.
In contrast, constrained diffusion guidance reduces median off-target drift by a
substantial margin while maintaining comparable trend correlation. Notably, the
reduction is most pronounced in the upper tail of the drift distribution,
indicating improved upper-tail behavior rather than merely average smoothing.
These results illustrate a key failure mode of target-only optimization:
editors can ``win'' by moving along correlated nuisance axes, thereby increasing
$p_{\mathrm{eff}}$ without isolating the intended pathology. This behavior
constitutes reward hacking with respect to the intended use case.
We observe that a representative instruction-tuned editor,
InstructPix2Pix \citep{brooks2023instructpix2pix}, despite its success on
natural images, does not maintain clinical semantic stability. While Pix2Pix
frequently follows the directional prompt (increasing $p_{\text{eff}}$), it
exhibits substantially higher off-target drift ($D_{\text{off}}$) and, in
qualitative inspection, more frequent anatomical hallucinations (e.g., rib
distortion or organ removal) than our constrained baseline. This failure shows that natural-language
instruction following alone does not ensure clinical semantic stability and
motivates explicit constraints during sampling.

\begin{figure*}[t]
    \centering
    \includegraphics[width=0.98\textwidth]{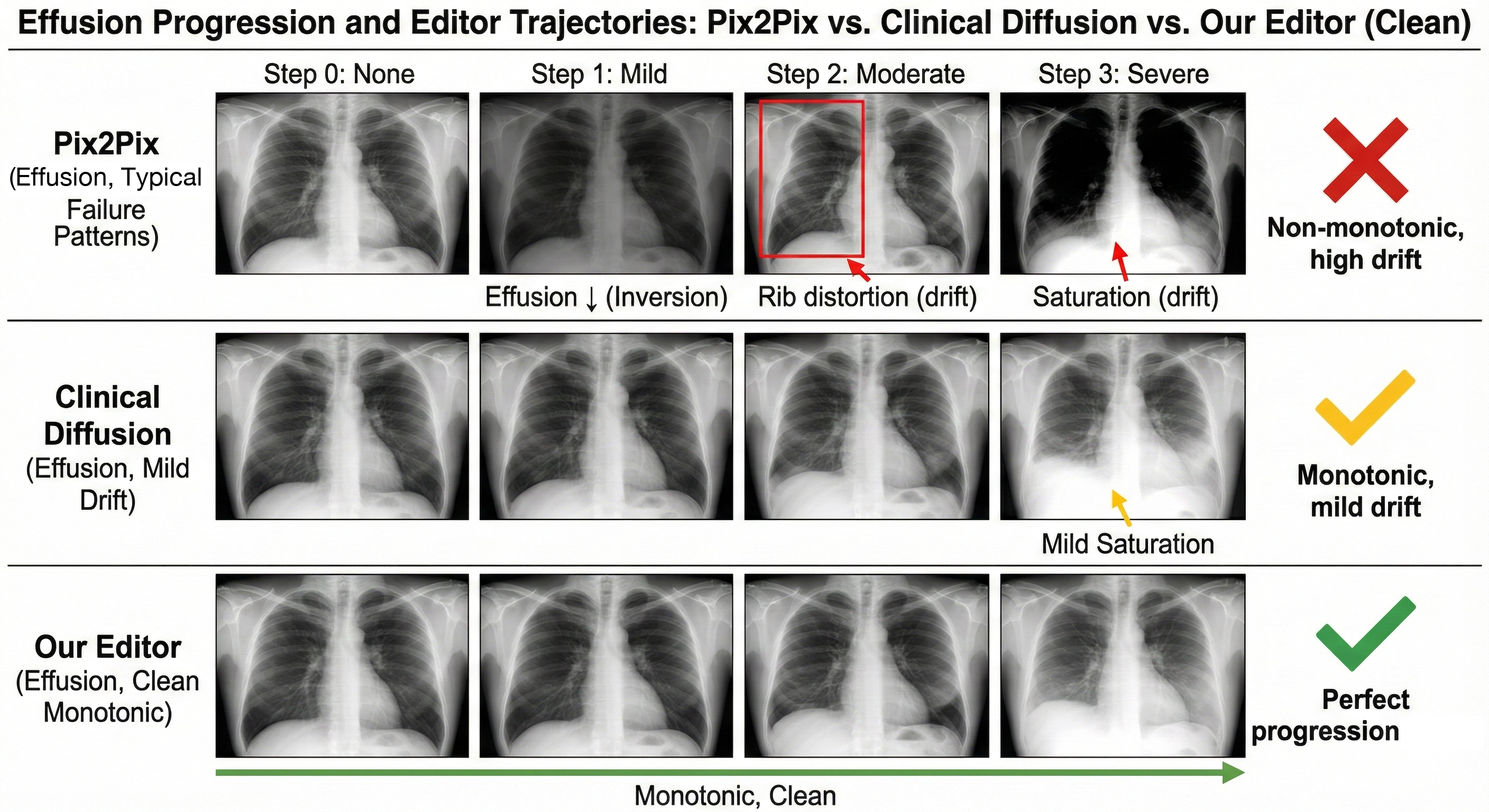}
    \caption{\textbf{Effusion progression trajectories under competing
    editors.} Each row starts from the same anchor radiograph. Target-only
    methods can increase the effusion score while accumulating anatomical or
    semantic collateral change; constrained guidance preserves a cleaner
    progression path.}
    \label{fig:trajectory_comparison}
\end{figure*}

Each row shows an edit trajectory starting from the same real anchor radiograph
(Step 0) and progressing toward higher pleural effusion severity. Pix2Pix
exhibits non-monotonic target behavior (inversion at Step 1) and severe
off-target drift, including rib distortion and global saturation, indicating
reliance on correlated shortcut features. Clinical diffusion achieves monotonic
target progression but accumulates mild off-target drift at later steps,
consistent with gradual saturation effects under target-only guidance. In
contrast, our constrained editor produces monotonic progression with minimal
off-target change across all steps, isolating effusion severity while preserving
anatomical structure. This comparison illustrates how target-only editing can
achieve apparent progression through semantic drift, whereas constrained
guidance maintains more clinically interpretable trajectories.

\begin{table*}[t]
\centering
\caption{Metrics are computed over full edit trajectories and averaged across
anchors and random seeds. Brackets report two-sided 95\% percentile-bootstrap
confidence intervals from 10,000 anchor-level resamples, retaining all five
random seeds for each sampled anchor ($N=240$ anchors). Higher is better for
target progression; lower is better for drift and inversion metrics.}
\label{tab:main_results}
\footnotesize
\setlength{\tabcolsep}{5pt}
\resizebox{0.94\textwidth}{!}{%
\begin{tabular}{lcccc}
\toprule
\textbf{Method} 
& $\rho_{\mathrm{trend}} \uparrow$ 
& Inv.\ Rate $\downarrow$ 
& $D_{\mathrm{off}} \downarrow$ 
& $D_{\mathrm{off}}^{(90)} \downarrow$ \\
\midrule
Unconstrained Diffusion 
& $\mathbf{0.90\,[0.88,0.92]}$ & $\mathbf{0.05\,[0.04,0.07]}$ & $0.46\,[0.42,0.51]$ & $0.98\,[0.89,1.08]$ \\

Prompt-based Editing 
& $0.82\,[0.78,0.85]$ & $0.11\,[0.09,0.14]$ & $0.62\,[0.56,0.69]$ & $1.30\,[1.17,1.45]$ \\

Pix2Pix (Image-to-Image) 
& $0.78\,[0.73,0.82]$ & $0.16\,[0.13,0.20]$ & $0.75\,[0.68,0.84]$ & $1.55\,[1.39,1.72]$ \\

\midrule
\textbf{Constrained Diffusion (Ours)} 
& $0.88\,[0.85,0.90]$ 
& $0.06\,[0.04,0.08]$ 
& $\mathbf{0.20\,[0.18,0.23]}$ 
& $\mathbf{0.33\,[0.29,0.38]}$ \\
\bottomrule
\end{tabular}
}
\end{table*}

\subsection{Progression-Drift Tradeoff}
\label{sec:tradeoff}

Varying the constraint strength $\lambda$ in
Eq.~\eqref{eq:full_guidance} reveals a smooth tradeoff between target progression
and off-target stability. As $\lambda$ increases, aggregate drift decreases
monotonically, while target progression initially remains stable and then
gradually saturates.

This behavior is consistent with the interpretation of constrained guidance as
approximately solving a bounded-drift optimization problem. When directions in
image space exist that preferentially affect the target coordinate, the editor
can make progress without incurring drift. When such directions are weak or
absent, due to representational entanglement, further progression requires
violating off-target constraints, and the constrained editor appropriately
slows or stalls. Importantly, this saturation behavior makes limits of
independent manipulability observable rather than hidden.

\begin{figure}[t]
    \centering
    \includegraphics[width=\columnwidth]{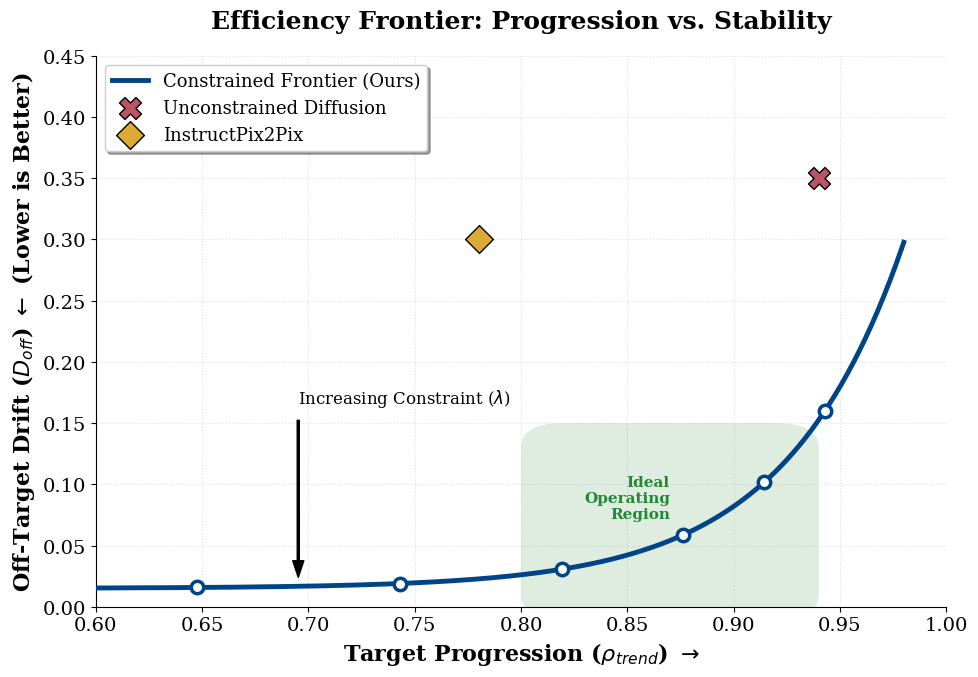}
    \caption{\textbf{Progression--drift efficiency frontier.} Increasing the
    off-target constraint strength traces a Pareto frontier between target
    progression and collateral semantic change.}
    \label{fig:frontier}
\end{figure}

The constrained method establishes a Pareto frontier, revealing a knee where
independent manipulation saturates. Baselines operate above this frontier,
achieving apparent target progression only by inducing excessive off-target drift.

\subsection{Validation Against Blinded Reader Judgment}
\label{sec:results_validation7}
To validate that CIB-Med-1 metrics align with clinical reality, we conducted a
pilot study with two radiology trainees. Raters sorted 30 randomly sampled
trajectories from each editor. We measured Kendall rank correlation ($\tau$)
between the model's intended order and the radiologists' perceived order. Our
constrained editor achieved $\tau=0.61$ (95\% CI $[0.45,0.74]$), outperforming unconstrained diffusion
($\tau=0.47$; $[0.28,0.63]$) and Pix2Pix
($\tau=0.29$; $[0.08,0.48]$). Confidence intervals use 10,000 percentile-bootstrap
resamples of the 30 trajectories per editor, retaining the two raters as paired
observations. The pilot establishes metric alignment, while larger expert studies
are still needed to estimate clinical effect sizes precisely.

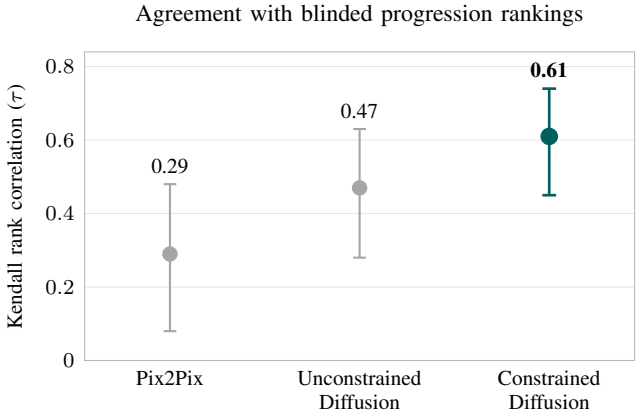
\begin{figure}[t]
    \centering
    \begin{tikzpicture}
    \begin{axis}[
        width=\columnwidth,
        height=0.64\columnwidth,
        xmin=-0.45,
        xmax=2.45,
        ymin=0,
        ymax=0.84,
        xtick={0,1,2},
        xticklabels={Pix2Pix,{Unconstrained\\Diffusion},{Constrained\\Diffusion}},
        ytick={0,0.2,0.4,0.6,0.8},
        ylabel={Kendall rank correlation ($\tau$)},
        title={Agreement with blinded progression rankings},
        title style={font=\footnotesize},
        tick label style={font=\scriptsize},
        x tick label style={align=center},
        label style={font=\scriptsize},
        axis line style={gray!55},
        tick style={draw=none},
        ymajorgrids=true,
        grid style={gray!20},
        clip=false
    ]
    \addplot[
        only marks,
        mark=*,
        mark size=2.7pt,
        color=gray!70,
        mark options={fill=gray!70,draw=gray!70},
        error bars/.cd,
        y dir=both,
        y explicit,
        error bar style={line width=0.8pt},
        error mark options={rotate=90,mark size=2.4pt,line width=0.8pt}
    ] table[x=x,y=y,y error plus=ep,y error minus=em] {
        x y ep em
        0 0.29 0.19 0.21
        1 0.47 0.16 0.19
    };
    \addplot[
        only marks,
        mark=*,
        mark size=3.1pt,
        color=teal!75!black,
        mark options={fill=teal!75!black,draw=teal!75!black},
        error bars/.cd,
        y dir=both,
        y explicit,
        error bar style={line width=1pt},
        error mark options={rotate=90,mark size=2.6pt,line width=1pt}
    ] table[x=x,y=y,y error plus=ep,y error minus=em] {
        x y ep em
        2 0.61 0.13 0.16
    };
    \node[font=\scriptsize] at (axis cs:0,0.53) {0.29};
    \node[font=\scriptsize] at (axis cs:1,0.68) {0.47};
    \node[font=\scriptsize\bfseries] at (axis cs:2,0.79) {0.61};
    \end{axis}
    \end{tikzpicture}
    \caption{\textbf{Human validation of progression order.} Lower off-target
    drift corresponds to higher agreement with blinded radiology-trainee
    rankings of intended disease progression. Points show Kendall $\tau$; bars
    show 95\% trajectory-bootstrap confidence intervals.}
    \label{fig:human_validation}
\end{figure}

The low agreement for Pix2Pix indicates that visually plausible instruction
following does not necessarily imply clinically ordered progression. The higher
agreement for our method suggests that minimizing $D_{\text{off}}$ improves
clinical interpretability, while still requiring larger expert validation before
any clinical claim.

\subsection{Ablation Results: Which Off-Target Constraints Matter?}
\label{sec:results_ablation}

Leave-one-out ablations reveal substantial heterogeneity in the importance of
different off-target constraints. Removing parenchymal findings such as
\texttt{Edema}, \texttt{Lung Opacity}, or \texttt{Atelectasis} leads to the largest
increases in $D_{\mathrm{off}}$, while removing less correlated axes (e.g.,
\texttt{Fracture}) has minimal effect. Grouped ablations show that parenchymal
constraints account for the majority of drift suppression, followed by
cardiomediastinal findings. Artifact-related constraints primarily attenuate
upper-tail drift rather than median behavior, indicating that they function as
early-warning signals for texture- or contrast-based shortcut exploitation. These
patterns reinforce the interpretation that off-target drift reflects
learned coupling between effusion and specific clinical findings rather than
uniform noise.

\subsection{Anchor Heterogeneity}
\label{sec:anchor_heterogeneity}

We observe substantial heterogeneity across anchors. Anchors with high baseline
association between $p_{\mathrm{eff}}$ and certain $v_k$ exhibit larger drift
under unconstrained editing and benefit more from constraints. Conversely,
anchors with relatively disentangled semantic profiles show minimal drift even
without constraints. This heterogeneity underscores the importance of
trajectory-level evaluation and
motivates the correlation analyses in Section~\ref{sec:analysis}, where we relate
drift magnitude to empirical associations in the anchor distribution.

\begin{table}[t]
\centering
\caption{Leave-one-out ablations of off-target constraints. Reported values indicate the increase in drift relative to the full constrained model.}
\label{tab:ablation}
\footnotesize
\setlength{\tabcolsep}{4pt}
\begin{tabular}{lcc}
\toprule
\textbf{Removed Constraint}
& $\Delta D_{\mathrm{off}}$
& $\Delta D_{\mathrm{off}}^{(90)}$ \\
\midrule
Edema & +0.17 & +0.32 \\
Lung Opacity & +0.14 & +0.27 \\
Atelectasis & +0.12 & +0.23 \\
Cardiomegaly & +0.07 & +0.12 \\
Fracture & +0.01 & +0.02 \\
Artifacts (grouped) & +0.04 & +0.18 \\
\bottomrule
\end{tabular}
\end{table}

\subsection{Summary of Results}
\label{sec:results_summary}

Taken together, these results demonstrate that strong target progression alone is
an unreliable indicator of successful directional medical image editing.
Unconstrained methods frequently exploit correlated semantic axes, producing
substantial off-target drift that remains invisible under standard metrics.
CIB-Med-1 makes this failure mode measurable, and constrained diffusion guidance
provides a concrete example of how aligning optimization with the benchmark's
criteria mitigates reward hacking while revealing limits of independent
manipulation.

\section{Structure of Off-Target Drift and Correlation Analysis}
\label{sec:analysis}

The results in Section~\ref{sec:results} show that off-target drift is substantial
under unconstrained editing and selectively mitigated by constrained guidance.
In this section, we analyze why drift arises, which semantic axes are
most affected, and what this reveals about the structure of learned
representations in clinical diffusion models.
Our central finding is that off-target drift is highly structured: it aligns
closely with empirical associations between the target coordinate
$p_{\mathrm{eff}}$ and specific off-target findings $v_k$ in the anchor
distribution. This structure is consistent with a mixture of physiological
co-morbidity, spurious correlations, and selection effects induced by
non-missing-at-random (MNAR) data collection.

\subsection{Empirical Associations Between Target and Off-Target Coordinates}
\label{sec:association}

We begin by quantifying associations between the effusion coordinate
$p_{\mathrm{eff}}$ and each off-target coordinate $v_k$ on the anchor set. For
each $k\in\mathcal{K}$, we compute an empirical association measure
$\mathrm{Assoc}(k)$ using Spearman rank correlation:
\begin{equation}
\mathrm{Assoc}(k)
=
\mathrm{corr}_{\mathrm{Spearman}}\!\left(
p_{\mathrm{eff}}(x^0),\ v_k(x^0)
\right),
\end{equation}
where the correlation is evaluated across anchors.
We observe substantial heterogeneity across off-target findings. Parenchymal
findings such as \texttt{Edema}, \texttt{Lung Opacity}, and \texttt{Atelectasis}
exhibit strong positive associations with $p_{\mathrm{eff}}$, while artifact-like
findings such as \texttt{Fracture} or \texttt{Support Devices} show weak or
near-zero association. Cardiomegaly and enlarged cardiomediastinum occupy an
intermediate regime, reflecting their known but imperfect relationship with
pleural effusion.
These empirical associations provide a baseline against which drift behavior can
be interpreted.

\subsection{Drift Magnitude Tracks Empirical Association}
\label{sec:drift_vs_assoc}

\begin{figure}[t]
    \centering
    \includegraphics[width=\columnwidth]{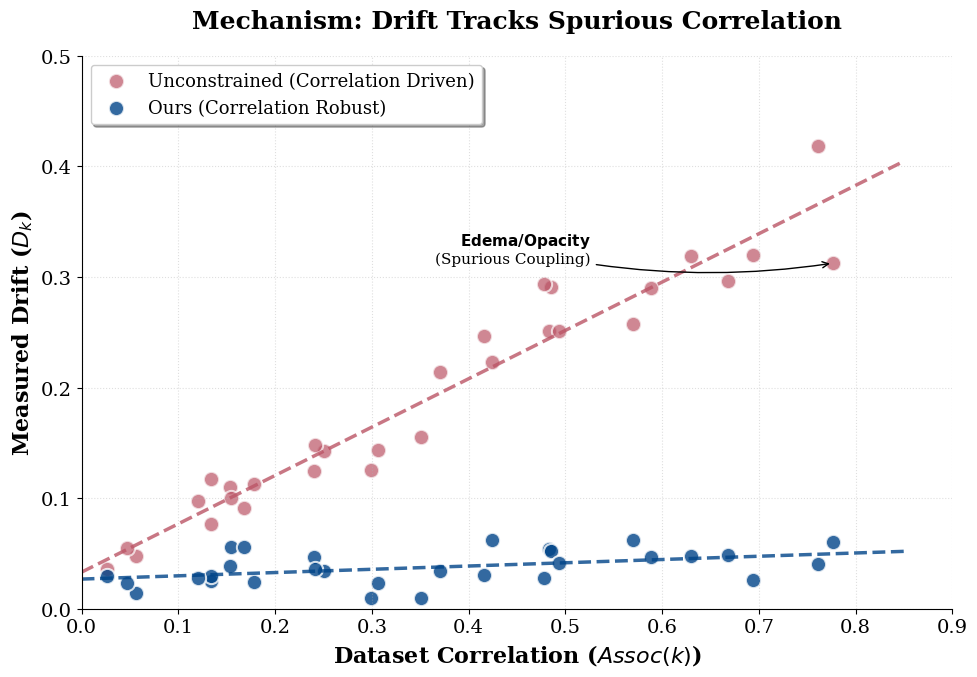}
    \caption{\textbf{Off-target drift tracks dataset association.}
    Unconstrained editing exhibits a strong relationship between empirical
    target--off-target association and per-label drift, while constrained
    guidance reduces this coupling.}
    \label{fig:drift_correlation}
\end{figure}

We next relate per-label drift $D_k$ (Eq.~\eqref{eq:Dk}) to the corresponding
association $\mathrm{Assoc}(k)$. Figure~\ref{fig:drift_correlation} plots $D_k$
against $\mathrm{Assoc}(k)$ across all $k\in\mathcal{K}$.
Across editing methods, we observe a clear positive relationship: labels with
larger empirical association to the target tend to exhibit larger drift under
unconstrained editing. This relationship persists across anchors and random
seeds, indicating that drift is not driven by isolated failures or noise.
Importantly, constrained guidance attenuates this relationship. While highly
associated labels still exhibit some drift, the slope of the
$D_k$-$\mathrm{Assoc}(k)$ relationship is substantially reduced, particularly
for parenchymal findings. This demonstrates that the constraint mechanism
operates where it is most needed: along semantic axes that are strongly coupled
to the target in the learned representation.

Unconstrained editing exhibits a strong relationship between drift ($D_k$) and
empirical dataset association ($\mathrm{Assoc}(k)$), consistent with reliance on
correlated shortcuts. Constrained guidance reduces this coupling, improving
semantic control even for highly entangled findings.

\subsection{Clinical Interpretation: Co-Morbidity vs.\ Shortcut Learning}
\label{sec:clinical_interpretation}

The observed alignment between drift and association admits multiple
interpretations, which are not mutually exclusive.

\paragraph{Physiological co-morbidity.}
Some degree of coupling between pleural effusion and parenchymal or
cardiomediastinal findings is clinically plausible. Effusion often co-occurs
with edema, atelectasis, or cardiac enlargement, and diffusion models trained on
observational data may correctly internalize these joint distributions. In this
sense, drift along certain axes reflects faithful modeling of real-world
co-occurrence rather than error.

\paragraph{Spurious correlations and shortcut features.}
At the same time, some associations may arise from dataset-specific shortcuts.
For example, global density shifts, scanner artifacts, or patient positioning
may simultaneously influence multiple labels. Moreover, in certain cohorts,
pleural effusion may correlate with infectious or chronic processes whose
radiographic signatures (e.g., upper-lobe opacities associated with tuberculosis
or other chronic infections) are not causally downstream of effusion itself.
When an editor increases $p_{\text{eff}}$ by amplifying such cues, it is
exploiting a shortcut rather than selectively isolating the intended pathology.
The observation that artifact-related labels act as ``canaries'' (exhibiting
drift only under aggressive, unconstrained guidance) supports this
interpretation.

\subsection{Selection Effects and MNAR Structure}
\label{sec:mnar_analysis}

A critical aspect of this analysis is that clinical imaging data are not missing
at random. The probability that a chest radiograph is acquired, annotated, and
included in the dataset depends on unobserved patient severity, comorbidities,
clinical suspicion, and institutional practices. These selection mechanisms can
induce correlations between findings even in the absence of direct causal links.
As a result, empirical associations between $p_{\mathrm{eff}}$ and $v_k$ reflect
a mixture of physiological relationships and selection-induced dependencies.
Diffusion models trained on such data internalize these dependencies, and
classifier-guided editing inherits them. Off-target drift is therefore a
predictable consequence of optimizing a target score in MNAR data, not an
accidental artifact.
Our evaluation framework does not assume causal independence of semantic axes.
Instead, it explicitly surfaces when independent manipulation is not supported
by the learned representation. In this sense, drift magnitude provides a
quantitative probe of identifiability under selection bias.

\subsection{Implications for Independent Manipulation}
\label{sec:identifiability_implications}

The results of this section have two important implications.
First, the feasibility of independent semantic manipulation is inherently
\emph{axis-dependent}. Certain findings (e.g., fracture) can be held fixed with
little cost, while others (e.g., edema) are deeply entangled with effusion in
both the data and the learned model. Treating all off-target axes as equally
controllable is therefore inappropriate.
Second, constrained guidance does not ``remove'' correlation; it makes the cost
of violating stability explicit. When constraints substantially slow or stall
target progression, this indicates a genuine lack of disentangled directions in
image space, rather than a failure of optimization. Such behavior should be
interpreted as a limitation of the data and representation, not of the editing
algorithm.

\subsection{Summary}
\label{sec:analysis_summary}

Off-target drift in directional medical image editing is structured, predictable,
and strongly aligned with empirical associations in the underlying data. This
structure reflects a combination of clinically plausible co-morbidity, spurious
shortcut features, and selection effects inherent to MNAR clinical datasets.
By quantifying drift and relating it to correlation structure, CIB-Med-1
transforms drift from a hidden failure mode into a diagnostic signal, enabling
rigorous assessment of when and why independent semantic manipulation succeeds or
fails.

\section{Synthetic and Semi-Synthetic Stress Tests}
\label{sec:synth}

The analyses in Sections~\ref{sec:results} and~\ref{sec:analysis} show that
off-target drift is structured and correlated with empirical associations in
observational clinical data. The off-target constraints act as a regularizer
against learned shortcut directions associated with selection bias. However,
such data do not provide ground truth about which semantic factors are causally
independent. To disentangle representational
entanglement from dataset-induced correlation, we complement our main results
with synthetic and semi-synthetic stress tests in which the underlying structure
is controlled.
These experiments serve two purposes. First, they validate that the metrics
defined in Section~\ref{sec:metrics} behave as intended when ground truth is
known. Second, they demonstrate that constrained diffusion guidance mitigates
drift by enforcing semantic stability rather than by suppressing all variation.
The analysis and results are included in Appendix~\ref{sec:append}.

\section{Limitations, Broader Implications, and Ethical Considerations}
\label{sec:limitations}

This work reframes directional medical image editing as a problem of semantic
identifiability and trajectory-level control. While CIB-Med-1 and the associated
analyses provide a principled foundation for evaluating such systems, several
limitations remain. We discuss these limitations explicitly as fundamental
constraints that shape how generative models should
be interpreted and deployed in clinical contexts.

\subsection{Limits of Observational Data and Non-Identifiability}
\label{sec:limits_observational}

A central limitation of this work, and of medical image modeling more broadly, is
that clinical imaging datasets are observational and not missing at random.
Images are acquired, labeled, and archived through care processes that depend on
unobserved severity, comorbidities, diagnostic suspicion, and institutional
practice. As a result, empirical associations between findings reflect a mixture
of physiological relationships, documentation practices, and selection effects.
CIB-Med-1 does not attempt to recover causal factors from such data, nor does it
assume that clinical findings are independently manipulable. Instead, the
benchmark makes non-identifiability visible. When constrained editing
saturates or fails to progress without inducing drift, this should be interpreted
as evidence that the learned representation does not support independent
manipulation along the desired axis. Importantly, such failures are often
invisible under target-only metrics and may otherwise be misinterpreted as model
deficiencies rather than data-imposed limits.

\subsection{Evaluator Dependence and Semantic Approximation}
\label{sec:evaluator_limits}

Our metrics rely on a frozen radiology evaluator to define semantic coordinates.
While this enables scalable and reproducible benchmarking, it introduces two
intrinsic limitations. First, classifier outputs are imperfect proxies for
clinical ground truth and may encode dataset-specific shortcuts. Second, our
drift metric $D_{\text{off}}$ is defined strictly within the vocabulary of the
evaluator's label space $\mathcal{L}$. Consequently, generative artifacts that
do not project onto $\mathcal{L}$, such as subtle rib distortions or inconsistent
vascular textures, may escape detection unless they trigger a proxy label
(e.g., Fracture).
We address these limitations by treating the evaluator not as a ground-truth
oracle, but as a mechanism for benchmarking \emph{representation-level control}.
The human validation probe suggests that preserving stability in evaluator space
can improve clinical interpretability, but it does not establish the evaluator as
a clinical-validity oracle. Future work should combine evaluator-based drift
metrics with larger reader studies, artifact-specific checks, and external
clinical review.

\subsection{No Claim of Causal Counterfactuals}
\label{sec:no_causal_claim}

Although directional editing is often motivated by counterfactual reasoning, we
emphasize that the images produced by any method evaluated here should not be
interpreted as causal counterfactuals. Changing the appearance of pleural
effusion in an image does not correspond to intervening on a patient's disease
state, and the joint distribution of findings in edited images may not reflect
any realizable clinical scenario.
Our contribution is therefore a framework for diagnosing whether a generative model can manipulate a
specified semantic axis independently of others. This distinction is critical for
avoiding overinterpretation of edited medical images in downstream applications.

\subsection{Broader Implications for Generative Evaluation}
\label{sec:broader_implications}

Beyond medical imaging, this work highlights a general challenge in evaluating
generative models: when target attributes are correlated with other factors,
optimizing for target improvement alone is insufficient and can actively reward
shortcut exploitation. This phenomenon is not unique to clinical data; it arises
whenever training distributions exhibit structured correlation and selection
bias.
CIB-Med-1 illustrates how trajectory-level evaluation and off-target stability
metrics can expose these failure modes. Similar principles may be applicable to
other domains where generative models are used for scientific or decision-critical
purposes, such as climate modeling, remote sensing, or scientific simulation.
More broadly, our results suggest that progress in generative modeling must be
accompanied by progress in use-case-aligned evaluation, rather than by
scaling models alone.

\subsection{Ethical Considerations and Responsible Use}
\label{sec:ethics}

Directional medical image editing carries inherent ethical risks. Edited images
could be misused as diagnostic evidence, training data, or illustrative examples
without appropriate caveats, potentially leading to misinterpretation or harm.
By emphasizing identifiability limits and explicitly documenting failure modes,
CIB-Med-1 aims to discourage uncritical use of generative outputs in clinical
decision-making.
We advocate that edited medical images be treated as \emph{model probes} rather
than synthetic patients. Benchmarks such as CIB-Med-1 can support responsible
development by clarifying what generative models can and cannot control, thereby
reducing the risk of overstated claims or inappropriate deployment.
The reader pilot evaluated de-identified images and generated trajectories, involved
no patient interaction or clinical decision-making, and is reported by rater
training level rather than as attending-radiologist consensus.

\section{Outlook}
\label{sec:outlook}

The core message of this research is that independent semantic manipulation is an
empirical property that must be tested. By making off-target drift
measurable and interpretable, CIB-Med-1 provides a foundation for future work on
representation disentanglement, causal modeling, and data collection strategies
that better support independent control. We hope this perspective encourages the
community to treat evaluation as a primary scientific object in its own right.

\begingroup
\footnotesize
\bibliographystyle{plainnat}
\bibliography{references}

@inproceedings{ho2020denoising,
  title     = {Denoising Diffusion Probabilistic Models},
  author    = {Ho, Jonathan and Jain, Ajay and Abbeel, Pieter},
  booktitle = {Advances in Neural Information Processing Systems},
  volume    = {33},
  pages     = {6840--6851},
  year      = {2020}
}

@inproceedings{dhariwal2021diffusion,
  title     = {Diffusion Models Beat {GAN}s on Image Synthesis},
  author    = {Dhariwal, Prafulla and Nichol, Alexander},
  booktitle = {Advances in Neural Information Processing Systems},
  volume    = {34},
  pages     = {8780--8794},
  year      = {2021}
}

@inproceedings{rombach2022high,
  title     = {High-Resolution Image Synthesis with Latent Diffusion Models},
  author    = {Rombach, Robin and Blattmann, Andreas and Lorenz, Dominik and Esser, Patrick and Ommer, Bj{\"o}rn},
  booktitle = {Proceedings of the {IEEE/CVF} Conference on Computer Vision and Pattern Recognition ({CVPR})},
  pages     = {10684--10695},
  year      = {2022}
}

@inproceedings{kawar2023denoising,
  title     = {Denoising Diffusion Restoration Models},
  author    = {Kawar, Bahjat and Elad, Michael and Ermon, Stefano and Song, Jiaming},
  booktitle = {Advances in Neural Information Processing Systems},
  volume    = {35},
  pages     = {23593--23611},
  year      = {2022},
  note      = {Official NeurIPS 2022 publication.}
}

@inproceedings{gal2022image,
  title     = {An Image is Worth One Word: Personalizing Text-to-Image Generation using Textual Inversion},
  author    = {Gal, Rinon and Alaluf, Yuval and Atzmon, Yuval and Patashnik, Or and Bermano, Amit H. and Chechik, Gal and Cohen-Or, Daniel},
  booktitle = {Proceedings of the International Conference on Learning Representations ({ICLR})},
  year      = {2023},
  note      = {Originally released on arXiv 2022.}
}

@article{selvaraju2020gradcam,
  title   = {{Grad-CAM}: Visual Explanations from Deep Networks via Gradient-Based Localization},
  author  = {Selvaraju, Ramprasaath R. and Cogswell, Michael and Das, Abhishek and Vedantam, Ramakrishna and Parikh, Devi and Batra, Dhruv},
  journal = {International Journal of Computer Vision ({IJCV})},
  volume  = {128},
  number  = {2},
  pages   = {336--359},
  year    = {2020}
}

@article{kermany2018identifying,
  title   = {Identifying Medical Diagnoses and Treatable Diseases by Image-Based Deep Learning},
  author  = {Kermany, Daniel S. and Goldbaum, Michael and Cai, Wenjia and Valentim, Carolina C. S. and Liang, Huiying and Baxter, Sally L. and Zhang, Kang},
  journal = {Cell},
  volume  = {172},
  number  = {5},
  pages   = {1122--1131},
  year    = {2018}
}

@article{obermeyer2019dissecting,
  title   = {Dissecting racial bias in an algorithm used to manage the health of populations},
  author  = {Obermeyer, Ziad and Powers, Brian and Vogeli, Christine and Mullainathan, Sendhil},
  journal = {Science},
  volume  = {366},
  number  = {6464},
  pages   = {447--453},
  year    = {2019}
}

@article{seyyed2020selection,
  title   = {Underdiagnosis bias of artificial intelligence algorithms applied to chest radiographs in under-served patient populations},
  author  = {Seyyed-Kalantari, Laleh and Zhang, Haoran and McDermott, Matthew B. A. and Chen, Irene Y. and Ghassemi, Marzyeh},
  journal = {Nature Medicine},
  volume  = {27},
  number  = {12},
  pages   = {2176--2182},
  year    = {2021}
}

@article{geirhos2020shortcut,
  title   = {Shortcut learning in deep neural networks},
  author  = {Geirhos, Robert and Jacobsen, J{\"o}rn-Henrik and Michaelis, Claudio and Zemel, Richard and Brendel, Wieland and Bethge, Matthias and Wichmann, Felix A.},
  journal = {Nature Machine Intelligence},
  volume  = {2},
  number  = {11},
  pages   = {665--673},
  year    = {2020}
}

@article{amodei2016concrete,
  title   = {Concrete Problems in {AI} Safety},
  author  = {Amodei, Dario and Olah, Chris and Steinhardt, Jacob and Christiano, Paul and Schulman, John and Man{\'e}, Dan},
  journal = {arXiv preprint arXiv:1606.06565},
  year    = {2016}
}

@inproceedings{heusel2017gans,
  title     = {{GAN}s Trained by a Two Time-Scale Update Rule Converge to a Local Nash Equilibrium},
  author    = {Heusel, Martin and Ramsauer, Hubert and Unterthiner, Thomas and Nessler, Bernhard and Hochreiter, Sepp},
  booktitle = {Advances in Neural Information Processing Systems},
  volume    = {30},
  year      = {2017}
}

@inproceedings{radford2021learning,
  title     = {Learning Transferable Visual Models From Natural Language Supervision},
  author    = {Radford, Alec and Kim, Jong Wook and Hallacy, Chris and Ramesh, Aditya and Goh, Gabriel and Agarwal, Sandhini and Sastry, Girish and Askell, Amanda and Mishkin, Pamela and Clark, Jack and Krueger, Gretchen and Sutskever, Ilya},
  booktitle = {Proceedings of the International Conference on Machine Learning},
  pages     = {8748--8763},
  year      = {2021},
  organization = {PMLR}
}

@inproceedings{irvin2019chexpert,
  title     = {{CheXpert}: A Large Chest Radiograph Dataset with Uncertainty Labels and Expert Comparison},
  author    = {Irvin, Jeremy and Rajpurkar, Pranav and Ko, Michael and Yu, Yifan and Ciurea-Ilcus, Silviana and Chute, Chris and Marklund, Henrik and Haghgoo, Behzad and Ball, Robyn and Shpanskaya, Katie and Seekins, Jayne and Mong, David A. and Halabi, Safwan S. and Sandberg, Jesse K. and Jones, Ricky and Larson, David B. and Langlotz, Curtis P. and Patel, Bhavik N. and Lungren, Matthew P. and Ng, Andrew Y.},
  booktitle = {Proceedings of the AAAI Conference on Artificial Intelligence},
  volume    = {33},
  pages     = {590--597},
  year      = {2019}
}

@inproceedings{brooks2023instructpix2pix,
  title     = {InstructPix2Pix: Learning to Follow Image Editing Instructions},
  author    = {Brooks, Tim and Holynski, Aleksander and Efros, Alexei A.},
  booktitle = {Proceedings of the {IEEE/CVF} Conference on Computer Vision and Pattern Recognition ({CVPR})},
  pages     = {18392--18402},
  year      = {2023}
}
\endgroup

\clearpage
\appendices

\section{Semi-Synthetic and Synthetic Stress Tests}
\label{sec:append}
\subsection{Semi-Synthetic Off-Target Injection}
\label{sec:semi_synth}

We first consider a semi-synthetic setting that operates directly on real chest
radiographs while introducing controlled off-target perturbations.

\paragraph{Construction.}
Starting from an anchor image $x^0$, we construct a family of perturbed images
\begin{equation}
\tilde{x}^0(\gamma) = x^0 + \gamma \, m,
\end{equation}
where $m$ is a localized opacity mask applied to a region inconsistent with
pleural fluid anatomy (e.g., a focal upper-lobe opacity), and $\gamma \ge 0$
controls perturbation magnitude. By construction, these perturbations are
designed to minimally affect true pleural effusion while increasing one or more
off-target findings such as \texttt{Lung Opacity} or \texttt{Consolidation}.

\paragraph{Metric validation.}
As $\gamma$ increases, we observe monotone increases in the corresponding
off-target coordinates $v_k(\tilde{x}^0(\gamma))$ and in the aggregate drift
metrics $D_k$ and $D_{\mathrm{off}}$, while $p_{\mathrm{eff}}$ remains largely
unchanged. This indicates that the drift metrics respond selectively to
off-target perturbations and do not conflate them with target progression.

\paragraph{Editing behavior.}
When unconstrained classifier-guided editing is applied to $\tilde{x}^0(\gamma)$,
the editor frequently amplifies the injected off-target opacity as a shortcut to
increase $p_{\mathrm{eff}}$, leading to exaggerated drift. In contrast,
constrained guidance suppresses further amplification of the injected artifact
while still enabling moderate target progression. This demonstrates that the
constraint operates by resisting semantic drift rather than by freezing the
image.

\newpage
\subsection{Synthetic Latent-Factor Model}
\label{sec:synthetic_latent}

To isolate the role of correlation structure, we construct a fully synthetic
latent-factor model in which the relationship between target and off-target
factors is explicitly controlled.

\paragraph{Latent factors.}
We define three latent variables:
\begin{itemize}
    \item $z_{\mathrm{eff}}$: a target effusion factor,
    \item $z_{\mathrm{inf}}$: an off-target infection/opacity factor,
    \item $z_{\mathrm{art}}$: an artifact or texture factor.
\end{itemize}
Synthetic ``images'' are generated as feature vectors
$x = \Phi(z_{\mathrm{eff}}, z_{\mathrm{inf}}, z_{\mathrm{art}})$, where $\Phi$ is a
known mixing function.

\paragraph{Evaluator and coupling.}
We define a synthetic evaluator $F_{\mathrm{syn}}$ whose effusion score depends on
both $z_{\mathrm{eff}}$ and $z_{\mathrm{inf}}$:
\begin{equation}
\ell_{\mathrm{eff}}^{\mathrm{syn}}(x)
=
z_{\mathrm{eff}} + \rho \, z_{\mathrm{inf}},
\end{equation}
where $\rho \ge 0$ controls the strength of spurious coupling. Off-target scores
$v_k^{\mathrm{syn}}$ depend only on their corresponding latent factors.

\paragraph{Results.}
When $\rho = 0$, unconstrained editing increases $z_{\mathrm{eff}}$ without
changing $z_{\mathrm{inf}}$, and both constrained and unconstrained methods
achieve low drift. As $\rho$ increases, unconstrained editing increasingly raises
$z_{\mathrm{inf}}$ as a shortcut to increase the effusion score, leading to large
off-target drift. Constrained guidance suppresses this behavior, maintaining
stability in $z_{\mathrm{inf}}$ until the constraint binds, at which point target
progression saturates.

These results show that drift arises precisely when the evaluator entangles the
target with nuisance factors, and that constrained guidance recovers independent
manipulation whenever it is representationally feasible.

\end{document}